# Cross-Enhancement Transform Two-Stream 3D ConvNets for Action Recognition

Dong Cao [†]
Institute of Cognitive Intelligence
DeepBlue Academy of Sciences
DeepBlue Technology
(Shanghai) Co., Ltd. No.369,
Weining Road, Shanghai, China
doocao@gmail.com

Lisha Xu[†]
Institute of Cognitive Intelligence
DeepBlue Academy of Sciences
DeepBlue Technology
(Shanghai) Co., Ltd. No.369,
Weining Road, Shanghai, China
xuls@deepblueai.com

Dongdong Zhang
DeepBlue Academy of Sciences
DeepBlue Technology
(Shanghai) Co., Ltd. No.369,
Weining Road, Shanghai, China
zhangdongdong@deepblueai.com

## ABSTRACT

Action recognition is an important research topic in computer vision. It is the basic work for visual understanding and has been applied in many fields. Since human actions can vary in different environments, it is difficult to infer actions in completely different states with a same structural model. For this case, we propose a Cross-Enhancement Transform Two-Stream 3D ConvNets algorithm, which considers the action distribution characteristics on the specific dataset. As a teaching model, stream with better performance in both streams is expected to assist in training another stream. In this way, the enhanced-trained stream and teacher stream are combined to infer actions. We implement experiments on the video datasets UCF-101, HMDB-51, and Kinetics-400, and the results confirm the effectiveness of our algorithm.

## 1 Introduction

In the field of visual understanding, action recognition is an important direction and play the role of cornerstone for further application. This paper studies the algorithm that improve the accuracy of action recognition when takes into account the action characteristic distribution.

There have been many action recognition algorithms proposed. Since convolutional neural networks (CNNs) perform well in the image recognition [3, 4, 5, 6], some works applied deep neural networks in action recognition [7, 8, 9]. Based on long-range temporal structure modeling, [24] provided temporal convolution network(TCN), and [25] proposed temporal segment network(TSN) for video-based action recognition. Two-stream models which combine the spatial information and motion information [14, 15, 17] are proposed then, improving accuracy significantly. However, in traditional two-stream approaches, dealing with optical flow occupies most time of the process [19, 20, 21, 22, 23]. Further, because optical flow computation is expensive and offline that cannot meet the demand of real-time application, 3D-ConvNet structure is proposed [10, 11, 12, 13]. 3D ConvNets extend the time domain in structure of CNN [3, 10] to extract spatial information and time information simultaneously, improving the training efficiency of the action recognition. Inflated 3D model(I3D) [1] inflates 2D kernel to 3D as well as combines RGB features and optical flow features to further improve system performance. Recently, Literature [16, 18] proposed a motion hallucination network, named MoNet. This novel network constructs the motion features based on appearance features, without calculating optical flow.

In this paper, we propose a novel algorithm for action recognition based on the Two-Stream algorithm, 3D ConvNets and transfer learning spirit, named cross-enhancement transform two-stream 3D ConvNets algorithm. The one of the two streams which has better performance is regarded as the teacher model to assist in training another stream.

## 2 Related Work

The literature [1] proposes an Inflated 3D ConvNets algorithm that significantly improves the accuracy of action recognition and has outstanding performance on typical datasets, including HMDB-51[28], UCF-101[29] and Kinetics-400[30] datasets. Implementing experiment with three setting of RGB, FLOW and RGB+FLOW, the performance of 3D-fused model (one I3D stream) is better than ConvNet+LSTM [9] and 3D-ConvNet [10, 11, 12, 13]. Two-Stream I3D algorithm is optimal, superior to the traditional Two-stream models [14, 15].

Analyzing the performance of the Two-Stream I3D algorithm further, we found that the performance of the optical flow stream is better than RGB stream on HMDB-51 and UCF-101 while the performance of RGB is better than optical flow on Kinetics-400. The performance of two-stream RGB+FLOW is better than single performance at both situations. The reason that the performance of RGB stream is better than optical flow on Kinetics-400 is due to more camera motion in the dataset, which increases the difficulty greatly in the feature extraction for motion information, resulting in the weaker performance in flow stream.

[†]Corresponding author

Our method differs from above method. The proposed cross-enhancement transform two-stream 3D ConvNets model transfers the features between RGB stream and flow stream, and adjusts the direction for different data characteristic distribution.

## 3 Cross-Enhancement Transformation for Two-Stream 3D ConvNets

The performance of whole I3D model is constrained due to the quite different performance of the RGB stream and optical flow stream [1]. We get inspiration from the literature [2, 31], which distills the motion feature from the trained optical flow stream to the RGB stream during the training of RGB stream, making the feature map of RGB stream containing both motion and appearance information.

Our model adds a mapping between the internal position of the two streams, promoting the stability of features as well as decreasing the training error of the augmented RGB stream [2]. Further, taking the dynamic characteristics into account, we discussed two structures of the cross-enhanced transform two-stream 3D ConvNets model on datasets HMDB-51 and UCF-101, and Kinetics-400.

### 3.1 Multiple-Enhanced-RGB Two Stream 3D ConvNets

Observing the performance of I3D algorithm [1] on HMDB-51 and UCF-101, we found that the accuracy of the optical flow stream is significantly higher than the RGB stream. On this premise, we reconstruct the 3D convNets to optimize the training phase and call it Multiple-Enhanced-RGB Two Stream 3D ConvNets.

**Training phase** Since the optical flow performance of the two-stream model is better on HMDB-51 and UCF-101, we first train the optical flow stream separately. After the optical flow stream training is completed, the flow stream model is required to be frozen, and all the parameters are fixed. Then, inspired by the generalized distillation [2, 27] we realize the transmission of the motion information extracted from optical flow stream to the RGB stream, which improves the learning efficiency of the RGB stream during the training phase.

Specifically, we establish two direct bridge mappings from the optical flow stream to the RGB stream within the network to achieve the transmission of motion feature information. As shown in the Figure 1, we first select a connection from 'a' to 'i' as the first bridge mapping according to the experimental results. This mapping corresponds to the first part $L_{UH1}$ of the complete loss function. The second mapping is the connection UH2 at the output location of the 3D ConvNets model, corresponding to the second part $L_{UH2}$ of the loss function. Here, Mean Squared Error (MSE) between these two mappings are supposed to be minimized to distill information. $L_{UH3}$ is the common cross entropy loss function between the prediction and ground truth. The complete loss function of the multiple-enhanced-RGB stream model is defined as follows:

$$\begin{aligned} loss_{UH} &= \alpha_1 L_{UH1} + \beta_1 L_{UH2} + \gamma_1 L_{UH3} \\ &= \alpha_1 \parallel Feature1_{RGB} - Feature1_{flow} \parallel^2 \\ &\quad + \beta_1 \parallel FeatureUH_{RGB} - FeatureUH_{flow} \parallel^2 \\ &\quad + \gamma_1 CrossEntropy(P_{UH}, y) \end{aligned} \quad (1)$$

Where $L_{UH1}$, $L_{UH2}$, $L_{UH3}$ represent the three stages of the loss function of the model, and $\alpha_1$, $\beta_1$, $\gamma_1$ are weights of them. $P_{UH}$ is the final prediction probability of HMDB-51 or UCF-101, and $y$ is the ground truth label.

In the three parts of the loss function, the related weights $\alpha_1, \beta_1, \gamma_1$ play a role of adjustment, to strengthen or suppress the constraint effect of each component in the loss function. If $\alpha_1 \neq 0, \beta_1 \neq 0, \gamma_1 \neq 0$, the values of $\alpha_1$, $\beta_1$ adjust the teaching effect of the flow stream on the RGB stream during the training process. The greater the relative values, the more significant the effect. If $\alpha_1 = \beta_1 = 0$, $\gamma_1 \neq 0$, the model degenerates to a common two-stream network structure [1]. If $\alpha_1 = 0$, $\beta_1 \neq 0$, $\gamma_1 \neq 0$, the model degenerates into a model similar to MARS [2].

Finally, we input the output result of the trained RGB stream and the frozen flow stream to a fully connection layer to learn the parameters while optimize the cross-entropy loss function.

**Test phase** During test phase, input the data into the multiple-enhanced-RGB two stream 3D ConvNets to infer the category of actions.

### 3.2 Multiple-Enhanced-Flow Two Stream 3D ConvNets

On Kinetics-400, it can be found that the accuracy of the RGB stream is superior to the optical flow stream in terms of the I3D [1]. As mentioned before, the main reason is that there is more camera motion noise in the Kinetics dataset. It is different from HMDB-51 and UCF-101, because camera noises increase the difficulty of actual motion feature extraction in optical flow stream, resulting in weaker performance of flow stream. Considering this problem, we adjust the model, and call it Multiple-Enhanced-Flow Two Stream 3D ConvNets.

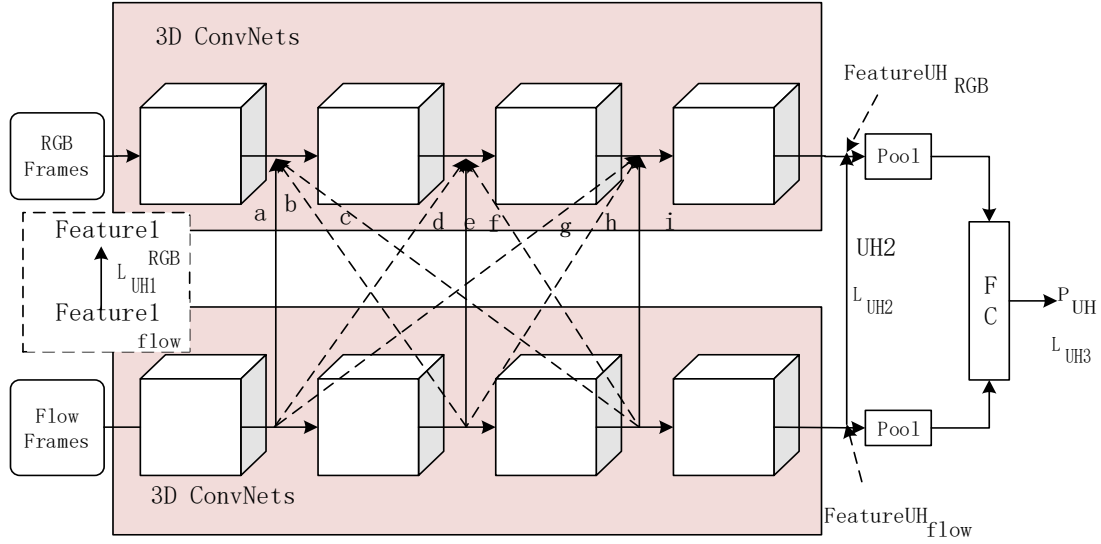

Figure 1: Multiple-Enhanced-RGB Two Stream 3D ConvNets on HMDB-51 and UCF-101.

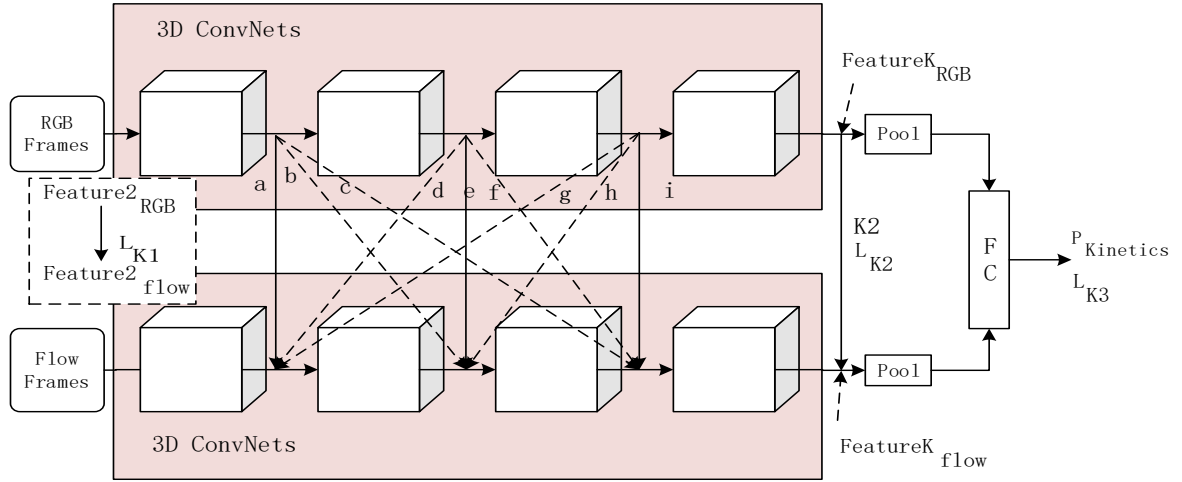

Figure 2: Multiple-Enhanced-Flow Two Stream 3D ConvNets for Kinetics-400.

**Training phase** Firstly, the optical flow stream does not participate in training while the RGB stream is trained. We froze the RGB stream and fix the trained parameters for following operation after the RGB stream finish training.

Next, similar to multiple-enhanced-RGB two stream 3D ConvNets, we establish two bridge mappings from RGB stream to optical flow stream. In Figure 2, one connection is chosen from 'a' to 'i', representing the transferring of the relatively specific RGB features. Another bridging mapping is the connection K2 at the rear of the 3D ConvNets model, representing the transferring of the abstract RGB features. Applying MSE to the distillation process realizes the appearance information transmission from the RGB stream to the optical flow stream. Though there are camera motion noises in the Kinetics-400 videos, the two bridge mappings can effectively play the role of motion stabilization by strengthening the appearance information, contributing to accuracy of optical flow stream. The complete loss function of the multiple-enhanced-flow stream model is defined as follows:

$$\begin{aligned} loss_K &= \alpha_2 L_{K1} + \beta_2 L_{K2} + \gamma_2 L_{K3} \\ &= \alpha_2 \parallel Feature2_{RGB} - Feature2_{flow} \parallel^2 \\ &\quad + \beta_2 \parallel FeatureK_{RGB} - FeatureK_{flow} \parallel^2 \\ &\quad + \gamma_2 CrossEntropy(P_{Kinetics}, y) \end{aligned} \quad (2)$$

Where $L_{K1}$, $L_{K2}$, $L_{K3}$ represent the three stages of the loss function of the model on Kinetics-400. $\alpha_2$, $\beta_2$, $\gamma_2$ are weights of them. $P_{Kinetics}$ is the final prediction probability of Kinetics-400, $y$ is the ground truth label.

Final training step is also the fully connection layer training to combine the RGB stream and flow stream.

**Test phase** After completing the above model's training, the bridge connections are removed and the complete two-stream model is retrained for inference.

## 4 Experiments

We evaluate our models on three datasets HMDB-51, UCF-101, and Kinetics-400, and compare the results with previous related algorithms on three settings of RGB, FLOW and RGB+FLOW.

### 4.1 Dataset

HMDB-51, UCF-101, and Kinetics-400 are baseline datasets in field of visual understanding. In Table1 and Table2, there are experimental settings and characteristic instructions of these datasets.

Table 1: The data setting of the experiments

| Dataset | Category | Total | Train | Test |
|---|---|---|---|---|
| HMDB-51 | 51 | 6766 | 5238 | 1528 |
| UCF-101 | 101 | 13320 | 9537 | 3783 |
| Kinetics-400 | 400 | 266440 | 246534 | 19906 |

Table 2: The characteristic induction of video datasets

| Dataset | Characteristic |
|---|---|
| HMDB-51 | Face action(smile, laugh); Face -object (smoke, eat); Body action(climb, dive); Body - object(brush hair, catch); Human - Human(hug, kiss) |
| UCF-101 | Body action(crawling); Body – object (brush teeth, haircut); Human – Human(lunges, frisbee catch); Playing instruments (drumming, playing piano); Playing Sports(basketball, bowling) |
| Kinetics-400 | Human(drawing, long jump); Human – Human(hugging, kissing); Human – object(brush painting, cleaning floor) |

### 4.2 Cross Enhancement Experiments

We select the front, medium and rear positions of the 3D ConvNet [1] model as the output and input positions for transferring specific features between the RGB stream and the optical flow stream. Thus, there are total 6 positions in the RGB stream and the optical flow stream, which are corresponding to nine bridging method as different models. Then we create another bridge connection at the output position of the two streams to realize the abstract feature transferring.

On HMDB-51 and UCF-101, the transmission is from the RGB stream to the flow stream. As for Kinetics-400, the feature information transfer in the opposite direction. Finally, we train this series of models separately to obtain the optimal model structure.

The test accuracy is shown in Table3. Our model has a significant improvement in related single stream experiments. Especially when there are some noises of camera movements or other motion information which does not belong to the action itself in the data, the multiple-enhanced-flow two stream 3D ConvNets algorithm is more suitable.

### 4.3 Implement details

We extract RGB frames at 25 fps for each video and compute optical flow using TV-L1 algorithm [21] by Opencv. During training, we use the SGD optimization method with a weight decay of 0.0005, momentum of 0.9, and an initial learning rate of 0.001 at 6 batchsize, 64 clip size and 224×224 frame size.

## 5 Conclusions

In this paper, we propose a novel model, named cross-enhancement transform two-stream 3D ConvNets. The stream with better performance in the two-stream model is regarded as the teacher model, in order to assist in training another stream. Specifically, the model enables to adjust the transmission direction according to characteristic distribution of data. The experimental results confirm the effectiveness of our algorithm on datasets HMDB-51, UCF-101, and Kinetics-400. And we will study better algorithms in future works.

Table 3: The test results of our model compared to the 3D-Fused model and two-stream I3D model.

| | HMDB-51 | | | UCF-101(split1) | | | Kinetics | | |
|---|---|---|---|---|---|---|---|---|---|
| | RGB | Flow | RGB+Flow | RGB | Flow | RGB+Flow | RGB | Flow | RGB+Flow |
| 3D-Fused | 49.2 | 55.5 | 56.8 | 83.2 | 85.8 | 89.3 | - | - | 67.2 |
| Two-Stream I3D | 49.8 | 61.9 | 66.4 | 84.5 | 90.6 | 93.4 | 71.1 | 63.4 | 74.2 |
| Ours | 58.7 | 61.9 | 66.6 | 86.1 | 90.6 | 93.6 | 71.1 | 65.3 | 74.5 |

## ACKNOWLEDGMENTS

We are very grateful to DeepBlue Technology (Shanghai) Co., Ltd. and DeepBlue Academy of Sciences for their support. Thanks to the support of the Equipment pre-research project (No. 31511060502). Thanks to Dr. ZhenDe Huang of the DeepBlue Academy of Sciences.